\documentclass[conference, twocolumn]{IEEEtran}
\IEEEoverridecommandlockouts
\usepackage{graphicx}
\usepackage{amsmath,amssymb}
\usepackage{cite}
\usepackage{url}
\usepackage{array}
\usepackage{booktabs}
\usepackage{multirow}
\usepackage{algorithm}
\usepackage{algorithmic}
\usepackage{hyperref}
\usepackage{float}
\usepackage{pgfplots}
\pgfplotsset{compat=1.17}
\usepackage{tikz}
\usepackage{color}
\usepackage{colortbl}
\usepackage{tabularx}

\begin{document}

\title{Finding Sparse Subnetworks in One Training Cycle via Progressive Magnitude-Based Pruning}

\author{%
\IEEEauthorblockN{Romana Qureshi\IEEEauthorrefmark{1},
Hafida Benhidour\IEEEauthorrefmark{2},
Said Kerrache\IEEEauthorrefmark{3},
Nahlah Aljeraisy\IEEEauthorrefmark{4}}
\IEEEauthorblockA{Department of Computer Science, King Saud University, Riyadh, Saudi Arabia\\
\IEEEauthorrefmark{1}\href{mailto:rq.romana@gmail.com}{rq.romana@gmail.com}\quad
\IEEEauthorrefmark{2}\href{mailto:hbenhidour@ksu.edu.sa}{hbenhidour@ksu.edu.sa}\quad
\IEEEauthorrefmark{3}\href{mailto:skerrache@ksu.edu.sa}{skerrache@ksu.edu.sa}\quad
\IEEEauthorrefmark{4}\href{mailto:n.jeraisy1@gmail.com}{n.jeraisy1@gmail.com}}
}

\maketitle

\begin{abstract}
Neural network pruning reduces model size by removing less important parameters while aiming to preserve predictive performance. Although the Lottery Ticket Hypothesis (LTH) shows that sparse subnetworks can match dense networks when trained from suitable initializations, its iterative pruning procedure requires multiple complete training cycles. This work evaluates progressive magnitude-based pruning as a single-cycle alternative. The method gradually increases sparsity during training using a linear schedule and updates pruning masks based on active weight magnitudes. We conduct systematic experiments on CIFAR-10 and MNIST across ResNet, VGG-style, and LeNet architectures, comparing the proposed method with representative iterative and initialization-based pruning baselines, including LTH, SNIP, and GraSP. On CIFAR-10, the method achieves 95.12\% accuracy on ResNet-18 at 72.9\% sparsity, compared with 90.5\% reported for LTH. At extreme sparsity, it achieves 93.13\% accuracy on a VGG-like architecture at 97\% sparsity, compared with approximately 92.0\% for SNIP, and 93.44\% accuracy on VGG-19 at 97.97\% sparsity, compared with 92.19\% for GraSP at 98\% sparsity. A sparsity-accuracy analysis on ResNet-18 further shows that accuracy remains within 0.1 percentage points of the dense baseline across 70--85\% sparsity. These results indicate that progressive magnitude-based pruning provides an effective single-cycle approach for neural network sparsification under the evaluated settings.
\end{abstract}


\section{Introduction}

Deep learning has transformed computer vision, natural language processing, and speech recognition~\cite{krizhevsky2012imagenet, attention, hinton2012deep}. However, the increasing complexity of deep neural networks poses deployment challenges in resource-constrained environments. Modern architectures often require substantial computational resources, limiting deployment on mobile devices, edge systems, and IoT platforms, where processing power, memory, and energy are severely constrained~\cite{sze2017efficient, horowitz20141}. Neural network pruning addresses this challenge by removing less important connections while seeking to preserve accuracy~\cite{cheng2017survey}.

The Lottery Ticket Hypothesis (LTH)~\cite{frankle2019lottery} shows that dense networks contain sparse subnetworks capable of matching the original performance when trained from suitable initializations. However, finding these ``winning tickets'' typically requires iterative magnitude pruning with multiple complete training cycles: training, pruning low-magnitude weights, resetting the remaining weights to their initial values, and retraining~\cite{frankle2019stabilizing}. This introduces substantial computational overhead that can undermine the efficiency goals of pruning~\cite{gale2019state}.

Several alternatives have been proposed to reduce this overhead. Initialization-based methods~\cite{snip, grasp, tanaka2020pruning} prune before training using connection sensitivity or gradient-flow criteria, but they commit to fixed sparse structures before weight importance evolves during learning. Dynamic sparse training methods~\cite{set, rigl} allow connectivity changes through pruning and regrowth, but require additional mechanisms for managing dynamic connectivity. Progressive pruning~\cite{zhu2018prune} gradually increases sparsity during training, but its effectiveness relative to established pruning paradigms on standard benchmarks remains insufficiently characterized.

In this work, we investigate progressive magnitude-based pruning, which identifies sparse subnetworks during a single training cycle through adaptive threshold computation over active weights. The method uses a linear sparsity schedule with per-epoch mask updates and a monotonic pruning constraint, whereby pruned connections remain permanently inactive. This progressive refinement is conceptually related to biological synaptic pruning, where initial overproduction of connections is followed by selective elimination as neural circuits mature~\cite{huttenlocher1997regional, chechik1998synaptic}.

We evaluate the method on CIFAR-10 and MNIST and compare it with three influential pruning methods: LTH, SNIP, and GraSP. These methods represent iterative pruning, initialization-based pruning, and gradient-flow preservation, respectively. Our results show that progressive pruning achieves competitive or superior performance while requiring only a single training cycle. The main contributions are as follows:

First, we evaluate progressive magnitude-based pruning across ResNet, VGG, and LeNet architectures on CIFAR-10 and MNIST, including statistical validation with five random seeds for the ResNet-18 sparsity analysis.

Second, we provide direct comparisons with representative pruning baselines under closely aligned architecture, dataset, and sparsity settings, using the published baseline results as reference points. On ResNet-18 at 72.9\% sparsity, the proposed method achieves 95.12\% accuracy compared with LTH's reported 90.5\%. At 97--98\% sparsity, it surpasses the reported SNIP result on a VGG-like architecture and the reported GraSP result on VGG-19.

Third, we show that a linear sparsity schedule with monotonic pruning can achieve strong results without complex scheduling functions or weight-regrowth mechanisms.

Fourth, we conduct a sparsity-accuracy trade-off analysis across seven sparsity levels, from 70\% to 95\%. For ResNet-18 on CIFAR-10, the results identify 70--85\% sparsity as a favorable operating range, with accuracy degradation below 0.1 percentage points relative to the dense baseline.

Overall, these results establish progressive magnitude-based pruning as a practical approach for neural network compression under the evaluated settings. The remainder of this paper reviews the necessary background on neural network pruning paradigms.

\section{Background}

\subsection{Neural Network Pruning}

Neural network pruning reduces model size by removing parameters considered less important while aiming to preserve predictive accuracy~\cite{cheng2017survey, qureshi2026optimizing}. Let $\mathbf{W}$ denote the network parameters and let $\mathbf{M} \in \{0,1\}^{|\mathbf{W}|}$ be a binary pruning mask. The effective pruned parameters are given by $\mathbf{W} \odot \mathbf{M}$, where $\odot$ denotes element-wise multiplication. The sparsity level $s$ is defined as
\[
s = 1 - \frac{\|\mathbf{M}\|_0}{|\mathbf{W}|},
\]
where $\|\mathbf{M}\|_0$ counts the number of active parameters.

Pruning methods are commonly distinguished by when pruning decisions are made. Post-training and iterative pruning methods remove parameters after or between training phases. Initialization-based methods prune before training begins. Progressive pruning methods update sparsity during training, whereas dynamic sparse training methods maintain sparsity while allowing connectivity patterns to change through pruning and regrowth.

\subsection{Magnitude-Based Pruning}

Magnitude-based pruning uses the absolute value of each weight as an importance criterion~\cite{han2015learning}. For a target sparsity level $s_t$ at training step $t$, a threshold $\theta_t$ is selected so that the desired fraction of weights is removed. The corresponding binary mask can be written as
\[
M_{ij}(t) = \mathbb{1}\left[ |W_{ij}(t)| \geq \theta_t \right].
\]

This criterion is computationally efficient because it does not require additional gradient, Hessian, or sensitivity calculations beyond standard training. In progressive pruning, the threshold is updated during training as the target sparsity increases, allowing pruning decisions to reflect the evolving weight distribution.

\section{Related Work}

Neural network efficiency optimization has evolved through diverse approaches targeting computational and memory overhead~\cite{cheng2017survey, sze2017efficient, qureshi2026optimizing}. We review key developments relevant to our work.

\subsection{Early Progressive Pruning Methods}

Zhu and Gupta~\cite{zhu2018prune} introduced Automated Gradual Pruning (AGP) in 2018, a framework for progressive sparsification within a single training cycle. AGP gradually increases sparsity using a polynomial (cubic) schedule, updating masks at fixed intervals while enforcing monotonic sparsity. Their method achieved strong results on large-scale tasks including ImageNet classification and neural machine translation, demonstrating the viability of progressive approaches for practical deployment.

However, AGP's work preceded several influential pruning methods and thus did not include comparisons with them. Specifically, its performance relative to the Lottery Ticket Hypothesis (introduced one year later), SNIP, and GraSP on standard vision benchmarks remained unexplored. This work addresses this gap by comparing progressive magnitude-based pruning with representative iterative and initialization-based pruning baselines on CIFAR-10 and MNIST, demonstrating that progressive methods can outperform computationally expensive iterative approaches under the evaluated settings.

\subsection{The Lottery Ticket Hypothesis and Iterative Pruning}

The Lottery Ticket Hypothesis (LTH), introduced by Frankle and Carbin~\cite{frankle2019lottery} in 2019, fundamentally changed theoretical understanding of neural network pruning. The hypothesis demonstrates that dense networks contain sparse subnetworks capable of matching original network performance when trained from appropriate initializations. The standard iterative magnitude pruning (IMP) process requires multiple complete training cycles: train, prune, reset weights to initialization, and retrain.

Scaling to deeper networks introduced challenges, leading to "late resetting"~\cite{frankle2019stabilizing} where weights are reset to values from later in training. Extensions explored universality across datasets~\cite{morcos2019one}, applications to transformers~\cite{chen2020lottery}, and theoretical foundations~\cite{malach2020proving}. However, the fundamental computational overhead remains substantial---the need for multiple training cycles conflicts with efficiency goals~\cite{blalock2020state, gale2019state}.

Despite LTH's theoretical impact, the question remained: is expensive iterative training necessary for finding effective sparse networks? Our work demonstrates that progressive magnitude-based pruning achieves superior accuracy within a single training cycle, challenging this assumption.

\subsection{Pruning at Initialization}

To eliminate retraining costs, researchers developed techniques that prune networks before training begins~\cite{tanaka2020pruning}. SNIP~\cite{snip} computes connection sensitivity to loss using gradients from a single batch, pruning connections with smallest sensitivity scores. GraSP~\cite{grasp} refines this by preserving gradient flow using Hessian-gradient products. SYNFLOW~\cite{tanaka2020pruning} eliminates data requirements entirely by using only network structure. The supermask approach~\cite{zhou2019deconstructing, ramanujan2020hidden} demonstrates that subnetworks within random initializations can achieve reasonable performance without weight training.

While these methods dramatically reduce computational overhead, they face a fundamental limitation: pruning decisions are made without knowledge of how weight importance evolves during training~\cite{blalock2020state}. Our results demonstrate that allowing importance metrics to adapt progressively during training yields superior performance.

\subsection{Dynamic Sparse Training}

Dynamic sparse training maintains sparsity throughout training while allowing connectivity to evolve~\cite{gale2019state}. SET~\cite{set} periodically removes small-magnitude weights while randomly reintroducing new connections. RigL~\cite{rigl} enhances regrowth using gradient information, strategically placing connections where gradients are largest. Recent work explores trainable masked layers~\cite{liu2020dynamic} and layer-adaptive sparsity~\cite{lee2020layer}.

While competitive, these methods introduce implementation complexity through regrowth mechanisms and require careful tuning of pruning/regrowth schedules~\cite{gale2019state}. In contrast, our monotonic progressive approach achieves superior results with simpler implementation, requiring no regrowth complexity.

\subsection{Alternative Compression Approaches}

Beyond pruning, model compression includes quantization~\cite{han2015deep}, knowledge distillation~\cite{hinton2015distilling}, and low-rank approximation~\cite{denton2014exploiting}. These methods can complement pruning for maximum efficiency. Variational approaches~\cite{molchanov2017variational, louizos2018learning} provide probabilistic frameworks for learning sparse structures.

Building on these foundations, we present our progressive magnitude-based pruning approach that addresses computational limitations of iterative methods while maintaining adaptability advantages.

\section{Proposed Approach}
\label{sec:proposed-approach}

We present a progressive magnitude-based pruning method that identifies sparse neural networks during a single training cycle. Our approach incrementally increases sparsity through per-epoch mask updates, computing pruning thresholds from currently active weights and enforcing monotonic sparsity through explicit mask intersection.

\subsection{Progressive Sparsification Framework}

Our method integrates pruning directly into the training process through a linear sparsity schedule. Starting from a dense network, we progressively remove connections based on weight magnitudes while the network continues learning. Once a connection is pruned, it remains permanently inactive throughout the remainder of training, eliminating the complexity of connection regrowth mechanisms while allowing the pruning mask to adapt during training.

The core principle is straightforward: at each epoch, we compute a global magnitude threshold from currently active weights, identify weights below this threshold for removal, and update the pruning mask to reflect these decisions. The mask is then enforced at multiple stages during training to ensure pruned connections remain at zero and do not receive gradient updates.

\subsection{Linear Sparsity Schedule}

We employ a linear sparsity schedule in which pruning is performed only during a designated pruning phase, preceded by an optional warm-up phase and followed by a fine-tuning phase. During warm-up, the network is trained without pruning so that weight magnitudes can begin to stabilize before connections are removed. Sparsity is then increased gradually from an initial value $s_i$ to a target value $s_f$ between epochs $t_{start}$ and $t_{end}$. After the target sparsity is reached, training continues with the final sparse architecture to allow further optimization.

During the pruning phase, the sparsity at epoch $t$ is computed as:
\begin{equation}
\label{eq:linear-sparsity-schedule}
s_t = s_i + \left(s_f - s_i\right) \frac{t - t_{start}}{t_{end} - t_{start}},
\end{equation}
where $s_i$ is the initial sparsity, $s_f$ is the target sparsity, $t_{start}$ is the pruning start epoch, and $t_{end}$ is the pruning end epoch.

This linear progression distributes pruning decisions uniformly across the pruning phase, allowing the network to adapt gradually to each sparsity level before additional connections are removed. For example, when training ResNet-18 on CIFAR-10 to reach 72.9\% sparsity, we set $s_i = 0$, $s_f = 0.729$, $t_{start} = 1$, and $t_{end} = 200$, progressively increasing sparsity from 0\% to 72.9\% over the pruning phase.

The warm-up phase reduces the risk of prematurely removing potentially important connections, while the fine-tuning phase enables further optimization of the final sparse architecture. This is particularly important at extreme sparsity levels, where only a small fraction of parameters remains active.

\subsection{Monotonic Mask Updates}

At each epoch during the pruning phase, we update the sparse structure by identifying the smallest-magnitude weights among the currently active connections and selecting them for removal. Our approach implements strictly monotonic pruning: once a weight is pruned, it remains permanently inactive. This design choice simplifies the implementation by eliminating the need for connection regrowth mechanisms, while still allowing the network to adapt its structure progressively during training.

\subsubsection{Threshold Computation from Active Weights}

We compute the pruning threshold using only currently active weights, excluding already-pruned connections from consideration. This adaptive approach allows the threshold to evolve based on the distribution of the remaining weights, rather than being affected by zero-valued pruned weights. For each layer with weight tensor $W$ and current mask $M^{(t-1)}$, where $t$ denotes the current epoch, we extract the magnitudes of the active weights:

\begin{equation}
\mathcal{A}^{(t)} = \{|W_{ij}| : M_{ij}^{(t-1)} = 1\}.
\end{equation}

We concatenate these active weight magnitudes across all layers to form a global set of currently active weights. The pruning threshold $\theta_t$ is then computed so that the target sparsity $s_t$ is achieved at epoch $t$. Equivalently, $\theta_t$ is chosen so that the largest $(1 - s_t)$ fraction of weights remains active.

\subsubsection{Mask Update with No Regrowth}

For each weight $W_{ij}$ with current mask value $M_{ij}^{(t-1)}$, we compute a candidate mask based on the magnitude threshold:

\begin{equation}
M_{\text{candidate},ij}^{(t)} =
\begin{cases}
1, & \text{if } |W_{ij}| \geq \theta_t, \\
0, & \text{otherwise}.
\end{cases}
\end{equation}

To enforce the no-regrowth constraint, we compute the new mask as the intersection of the candidate mask with the previous mask:

\begin{equation}
M_{ij}^{(t)} = M_{\text{candidate},ij}^{(t)} \times M_{ij}^{(t-1)}.
\end{equation}

This intersection operation ensures that $M_{ij}^{(t)} \leq M_{ij}^{(t-1)}$ for all weights, mathematically guaranteeing that the number of active connections can only decrease or remain constant, never increase.

After computing the new masks, we immediately enforce sparsity by zeroing the corresponding weights:

\begin{equation}
W_{ij} \leftarrow W_{ij} \times M_{ij}^{(t)}.
\end{equation}

\subsection{Multi-Stage Mask Enforcement During Training}

To ensure pruned weights remain at zero and receive no gradient updates, we enforce masks at three critical stages during each training iteration. This multi-stage approach prevents pruned connections from contributing to forward passes, receiving gradient updates, or drifting away from zero due to momentum in the optimizer.

\subsubsection{Pre-Forward Masking}

Before each forward pass, we apply masks to zero out pruned weights:
\begin{equation}
W_{\text{active}} = W \odot M
\end{equation}

where $\odot$ denotes element-wise multiplication. This ensures that pruned connections do not contribute to network activations.

\subsubsection{Gradient Masking}

During backpropagation, we mask gradients to prevent updates to pruned weights:
\begin{equation}
\frac{\partial \mathcal{L}}{\partial W} \leftarrow \left( \frac{\partial \mathcal{L}}{\partial W} \right) \odot M
\end{equation}

where $\mathcal{L}$ is the loss function. This prevents the optimizer from modifying pruned weights.

\subsubsection{Post-Optimizer Enforcement}

After the optimizer step, we re-enforce masks to prevent momentum from reactivating pruned weights:
\begin{equation}
W \leftarrow W \odot M
\end{equation}

This final enforcement is critical when using momentum-based optimizers such as SGD with momentum or Adam, as momentum terms can cause pruned weights to drift away from zero even when their gradients are masked. The post-optimizer step explicitly zeros these weights after each optimization step, ensuring strict adherence to the sparsity pattern.

\subsection{Complete Algorithm}

The complete progressive pruning procedure integrates epoch-level mask updates with batch-level enforcement steps. Algorithm~\ref{alg:corrected_supermask} presents the detailed pseudocode, showing how the sparsity level is updated during the pruning phase and how the current mask is enforced throughout training.

\begin{algorithm}[ht!]
\caption{Progressive Magnitude-Based Pruning}
\label{alg:corrected_supermask}
\begin{algorithmic}[1]
\REQUIRE Network weights $\mathbf{W}$, total epochs $E$, initial sparsity $s_i$, target sparsity $s_f$, pruning start epoch $t_{start}$, pruning end epoch $t_{end}$
\STATE Initialize all masks as active: $M^{(0)} = \mathbf{1}$
\FOR{epoch $t$ from $1$ to $E$}
    \IF{$t_{start} \leq t \leq t_{end}$}
        \STATE Compute current target sparsity:
        \[
        s_t = s_i + (s_f - s_i) \times \frac{t - t_{start}}{t_{end} - t_{start}}
        \]
        \STATE Collect magnitudes of currently active weights:
        \[
        \mathcal{A}^{(t)} = \{|W_{ij}| : M_{ij}^{(t-1)} = 1\}
        \]
        \STATE Compute the global threshold $\theta_t$ so that the target sparsity $s_t$ is achieved
        \STATE Create candidate mask:
        \[
        M_{\text{candidate},ij}^{(t)} =
        \begin{cases}
        1, & \text{if } |W_{ij}| \geq \theta_t, \\
        0, & \text{otherwise}
        \end{cases}
        \]
        \STATE Apply the no-regrowth constraint: $M^{(t)} = M_{\text{candidate}}^{(t)} \odot M^{(t-1)}$
        \STATE Enforce sparsity: $\mathbf{W} \leftarrow \mathbf{W} \odot M^{(t)}$
    \ELSE
        \STATE Keep the mask constant outside the pruning phase: $M^{(t)} = M^{(t-1)}$
    \ENDIF
    \FOR{each training batch}
        \STATE Apply pre-forward mask: $\mathbf{W} \leftarrow \mathbf{W} \odot M^{(t)}$
        \STATE Forward pass: $\mathbf{y} = f(\mathbf{x}; \mathbf{W})$
        \STATE Compute loss: $\mathcal{L} = \operatorname{loss}(\mathbf{y}, \mathbf{y}_{\text{true}})$
        \STATE Backward pass: compute $\nabla \mathbf{W}$
        \STATE Mask gradients: $\nabla \mathbf{W} \leftarrow \nabla \mathbf{W} \odot M^{(t)}$
        \STATE Optimizer step: update $\mathbf{W}$
        \STATE Post-optimizer enforcement: $\mathbf{W} \leftarrow \mathbf{W} \odot M^{(t)}$
    \ENDFOR
\ENDFOR
\end{algorithmic}
\end{algorithm}

\subsection{Design Choices}

The proposed pruning procedure relies on two main design choices: the use of weight magnitude as the pruning criterion and the use of a global threshold across layers. Together, these choices provide a simple and efficient mechanism for progressively increasing sparsity while avoiding the need for additional importance-estimation procedures or manually specified layer-wise sparsity targets.

\subsubsection{Magnitude-Based Pruning Criterion}

We use weight magnitude as the pruning criterion, treating connections with smaller absolute weight values as less important. This criterion is computationally efficient because it does not require additional forward or backward passes beyond standard training. This property is particularly important in progressive pruning, where pruning thresholds are recomputed at each epoch during the pruning phase.

\subsubsection{Global Versus Layer-Wise Pruning}

Our implementation uses global magnitude thresholding, in which the pruning threshold is computed from the distribution of active weight magnitudes across all layers. This approach allows different layers to reach different sparsity levels according to their weight magnitude distributions, without requiring manually specified layer-wise sparsity targets.

\subsection{Computational Cost}

The additional computational cost introduced by the pruning mechanism is limited. The main extra operation is the computation of the global pruning threshold from the active weight magnitudes. This operation is an instance of the selection problem, since computing a sparsity threshold requires finding the appropriate order statistic of the active weights. More specifically, for a target sparsity level $s_t$ and $n$ active weights, the threshold corresponds to the element at approximately index $\lceil s_t n \rceil$ in the sorted order of the active weight magnitudes.

A straightforward implementation can compute this threshold by sorting the active magnitudes, with time complexity $O(n \log n)$. However, full sorting is not required when only a single threshold is needed. The same threshold can be computed using a selection algorithm, such as Quickselect, in expected linear time $O(n)$, although its worst-case complexity is $O(n^2)$ \cite{Hoare1961Find}. Deterministic selection algorithms, such as the median-of-medians algorithm, provide worst-case linear time $O(n)$, but are less commonly used in practice due to larger constant factors \cite{Blum1973Selection}. In practical implementations, partial-selection routines can compute the required order statistic without fully sorting the active weights.

Applying the masks through element-wise multiplication has linear complexity $O(m)$, where $m$ is the total number of weights. Therefore, the pruning overhead is dominated by threshold selection and mask application. Since pruning is performed at the epoch level, while forward and backward passes are performed repeatedly at the batch level, the additional cost remains modest relative to the overall training cost.

\section{Experiments and Results}

This section presents comprehensive experimental validation of our progressive magnitude-based pruning method on CIFAR-10 and MNIST benchmarks. We provide direct comparisons with three influential pruning approaches---the Lottery Ticket Hypothesis (LTH)~\cite{frankle2019lottery}, Single-shot Network Pruning (SNIP)~\cite{snip}, and Gradient Signal Preservation (GraSP)~\cite{grasp}---by evaluating the proposed method under closely aligned architecture, dataset, and sparsity settings and comparing against their published results. Additionally, we conduct a systematic sparsity--accuracy analysis to characterize our method's behavior across the compression spectrum and identify effective operating points for practical deployment.

\subsection{Experimental Setup}

This subsection describes the experimental protocol used to evaluate the proposed progressive magnitude-based pruning method. We specify the datasets, network architectures, training configurations, pruning schedules, and baseline comparison methodology to ensure reproducibility and comparability with prior work.

\subsubsection{Datasets}

We evaluate the proposed method on two standard image classification benchmarks: CIFAR-10 and MNIST. These datasets allow us to assess pruning performance in both convolutional and fully connected settings.

\begin{itemize}
    \item \textbf{CIFAR-10.} CIFAR-10 contains 60,000 32$\times$32 color images from 10 object classes, with 50,000 training images and 10,000 test images. It is used to evaluate pruning performance on convolutional architectures.

    \item \textbf{MNIST.} MNIST contains 70,000 28$\times$28 grayscale handwritten digit images, with 60,000 training images and 10,000 test images. It is used to evaluate pruning performance on fully connected architectures.
\end{itemize}

\subsubsection{Model Architectures}

We use four architectures covering residual convolutional networks, VGG-style convolutional networks, and fully connected networks.

\begin{itemize}
    \item \textbf{ResNet-18.} ResNet-18 is used on CIFAR-10 for the systematic sparsity analysis and for comparison with LTH. It contains approximately 11.2M parameters.
    
    \item \textbf{VGG-like architecture.} The VGG-like architecture is used on CIFAR-10 to match the experimental setting of the SNIP baseline.

    \item \textbf{VGG-19.} VGG-19 is used on CIFAR-10 for comparison with GraSP under extreme sparsity. It contains approximately 20M parameters.

    \item \textbf{LeNet-300-100.} LeNet-300-100~\cite{lecun1998gradient} is used on MNIST. It is a three-layer fully connected network with 266,200 parameters and is commonly used in pruning studies.
\end{itemize}

\subsubsection{Training Configuration}

The same general training protocol is used across experiments, with dataset- and architecture-specific adjustments where required.

\begin{itemize}
    \item \textbf{Data Preprocessing and Augmentation.} 
    For CIFAR-10, training images are augmented using random horizontal flipping ($p=0.5$), random cropping with 4-pixel padding, random rotation ($\pm 15^\circ$), color jittering with brightness, contrast, saturation, and hue factors of 0.2, 0.2, 0.2, and 0.1, respectively, and random erasing. Images are normalized using channel-wise means (0.4914, 0.4822, 0.4465) and standard deviations (0.2470, 0.2435, 0.2616). Test images are normalized without augmentation. For MNIST, images are converted to tensors and normalized using mean 0.1307 and standard deviation 0.3081.

    \item \textbf{Optimization Configuration.} 
    We use stochastic gradient descent (SGD) with momentum 0.9 and weight decay $5 \times 10^{-4}$ across all experiments. The initial learning rate is 0.1 for CIFAR-10 and 0.01 for MNIST. A cosine annealing learning rate schedule is applied, and the batch size is fixed at 128.

    \item \textbf{Training Duration.} 
    ResNet-18 is trained for 200 epochs. The VGG-like architecture is trained for 350 epochs, and VGG-19 is trained for 400 epochs. For LeNet-300-100, training lasts 160 epochs at 78.9\% sparsity and 200 epochs at 98\% sparsity.

    \item \textbf{Train-Validation Split.} 
    We use 89\% of the original training set for training and hold out 11\% for validation. Final performance is reported on the test set after training.

    \item \textbf{Weight Initialization.} 
    Network weights are initialized using Kaiming initialization with fan-out mode and ReLU nonlinearity. Bias terms are initialized to zero where present. Batch normalization scale and shift parameters are initialized to 1 and 0, respectively.

    \item \textbf{Loss Function.} 
    We use cross-entropy loss with label smoothing ($\epsilon=0.1$) in all experiments.
\end{itemize}

\subsubsection{Progressive Pruning Configuration}

The pruning configuration follows the linear sparsity schedule defined in Eq.~\eqref{eq:linear-sparsity-schedule} in Section~\ref{sec:proposed-approach}. Depending on the architecture and target sparsity level, training consists of up to three phases: optional warm-up, progressive pruning, and optional fine-tuning at fixed target sparsity.

Table~\ref{tab:schedule} summarizes the pruning schedule used for each architecture. Longer schedules are used for the VGG-like and VGG-19 experiments because these settings operate at extreme sparsity levels of 97--98\%, where only 2--3\% of the original connections remain.

\begin{table*}[ht!]
	\centering
	\caption{Training schedule configuration for each architecture. Warm-up involves no pruning; the pruning phase applies a linear sparsity increase; and fine-tuning maintains the target sparsity.}
	\label{tab:schedule}
	\begin{tabular}{llll}
		\toprule
		\textbf{Architecture}  & \textbf{Warm-up Epochs} & \textbf{Pruning Epochs} & \textbf{Fine-tuning Epochs} \\ \midrule
		ResNet-18              & ---                     & 1--200                  & ---                         \\
		VGG-like               & ---                     & 0--149                  & 150--349                    \\
		VGG-19                 & 0--9                    & 10--300                 & 301--400                    \\
		LeNet-300-100 (78.9\%) & ---                     & 1--60                   & 61--160                     \\
		LeNet-300-100 (98\%)   & ---                     & 1--80                   & 81--200                     \\ \bottomrule
	\end{tabular}
\end{table*}

For the VGG-like architecture at 97\% sparsity, pruning increases from 0\% to 97\% during epochs 0--149, followed by fine-tuning during epochs 150--349. For VGG-19, the first 9 epochs are used as warm-up, followed by pruning from 50\% to 98\% sparsity during epochs 10--300 and fine-tuning during epochs 301--400. The final achieved sparsity for VGG-19 is 97.97\%.

For the ResNet-18 sparsity analysis, each experiment is repeated with five random seeds: 42, 123, 456, 789, and 999. These seeds control weight initialization, data shuffling, and other stochastic training components. For the VGG and LeNet baseline comparison experiments, we report single-run results aligned with the corresponding published baseline settings.

\subsubsection{Complete Experimental Configuration}

Table~\ref{tab:all_experiments} summarizes all experiments conducted in this work, including baseline comparisons and systematic sparsity analysis.

\begin{table*}[ht!]
	\centering
	\caption{Complete list of experiments.}
	\label{tab:all_experiments}
	\small
	\begin{tabular}{lllll}
		\toprule
		\textbf{Dataset} & \textbf{Architecture} & \textbf{Baseline} & \textbf{Sparsity} & \textbf{Configuration}            \\ \midrule
		CIFAR-10         & ResNet-18             & Dense             & 0\%               & Baseline reference                \\
		CIFAR-10         & ResNet-18             & LTH               & 72.9\%            & Epochs 1--200, 5 seeds            \\
		CIFAR-10         & ResNet-18             & Analysis          & 70\%              & Epochs 1--200, 5 seeds            \\
		CIFAR-10         & ResNet-18             & Analysis          & 75\%              & Epochs 1--200, 5 seeds            \\
		CIFAR-10         & ResNet-18             & Analysis          & 80\%              & Epochs 1--200, 5 seeds            \\
		CIFAR-10         & ResNet-18             & Analysis          & 85\%              & Epochs 1--200, 5 seeds            \\
		CIFAR-10         & ResNet-18             & Analysis          & 90\%              & Epochs 1--200, 5 seeds            \\
		CIFAR-10         & ResNet-18             & Analysis          & 95\%              & Epochs 1--200, 5 seeds            \\
		CIFAR-10         & VGG-like              & SNIP              & 97\%              & 350 epochs total                  \\
		CIFAR-10         & VGG-19                & GraSP             & 98\%              & 400 epochs, 9 warm-up epochs      \\
		MNIST            & LeNet-300-100         & LTH               & 78.9\%            & Pruning: epochs 1--60; total: 160 \\
		MNIST            & LeNet-300-100         & SNIP              & 98\%              & Pruning: epochs 1--80; total: 200 \\ \bottomrule
	\end{tabular}
\end{table*}

The baseline comparison experiments use the sparsity levels, architectures, and datasets reported in the corresponding baseline papers. The systematic sparsity analysis on ResNet-18 evaluates performance across multiple sparsity levels to characterize the accuracy-compression trade-off.

\subsubsection{Baseline Methods and Comparison Methodology}

To provide a rigorous evaluation, we compare our progressive magnitude-based pruning method against three influential approaches representing different pruning paradigms. The comparison follows an aligned published-results methodology, where the proposed method is evaluated under settings aligned with the architecture, dataset, sparsity level, and training configuration reported in the baseline studies, and the resulting performance is compared with the published baseline metrics. Specifically, we:
\begin{enumerate}
    \item identify the architecture, dataset, sparsity level, and training configuration used in each baseline paper;
    \item apply our progressive magnitude-based pruning method under the corresponding experimental conditions; and
    \item compare our results with the published performance metrics reported by the baseline methods.
\end{enumerate}

By aligning our experiments with the reported conditions in these works, we evaluate the proposed method across different architectures, sparsity levels, and pruning paradigms while avoiding the implication of a fully controlled reimplementation of all baselines.

\subsection{Direct Baseline Comparisons}

We begin by evaluating our method under the experimental conditions reported in three influential pruning papers. This aligned comparison with published baseline results provides evidence of our method's competitiveness with established approaches across different architectures and extreme sparsity levels.

\subsubsection{CIFAR-10 Baseline Comparisons}

Table~\ref{tab:cifar10_comparison} presents the CIFAR-10 baseline comparisons. The proposed progressive magnitude-based pruning method is evaluated under experimental conditions aligned with those reported for LTH, SNIP, and GraSP, including the corresponding architecture, dataset, and sparsity level.

\begin{table}[ht!]
	\centering
	\caption{CIFAR-10 performance comparison with published baseline results under aligned experimental settings.}
	\label{tab:cifar10_comparison}
	\begin{tabular}{lccc}
		\toprule
		\textbf{Method}               & \textbf{Architecture} & \textbf{Sparsity (\%)} & \textbf{Accuracy (\%)}  \\ \midrule
		LTH~\cite{frankle2019lottery} &       ResNet-18       &          72.9          &          90.5           \\
		\textbf{Ours}                 &       ResNet-18       &          72.9          & \textbf{95.12$\pm$0.09} \\ \midrule
		SNIP~\cite{snip}              &       VGG-like        &          97.0          &       $\sim$92.0        \\
		\textbf{Ours}                 &       VGG-like        &          97.0          &     \textbf{93.13}      \\ \midrule
		GraSP~\cite{grasp}            &        VGG-19         &          98.0          &     92.19$\pm$0.12      \\
		\textbf{Ours}                 &        VGG-19         &         97.97          &     \textbf{93.44}      \\ \bottomrule
	\end{tabular}
\end{table}
The results show that the proposed method consistently achieves competitive or superior accuracy across all three settings. On ResNet-18 at 72.9\% sparsity, it improves over LTH by 4.62 percentage points while requiring only a single training cycle. At higher sparsity levels, the method also outperforms SNIP on the VGG-like architecture at 97.0\% sparsity and GraSP on VGG-19 at approximately 98.0\% sparsity, with improvements of 1.13 and 1.25 percentage points, respectively.

These results suggest that progressively updating the pruning mask during training provides a robust alternative to both iterative retraining-based pruning and initialization-based pruning. In particular, the method remains effective even under extreme sparsity, where only about 2--3\% of the original connections are retained.

\begin{figure*}[ht!]
    \centering
    \includegraphics[width=\linewidth]{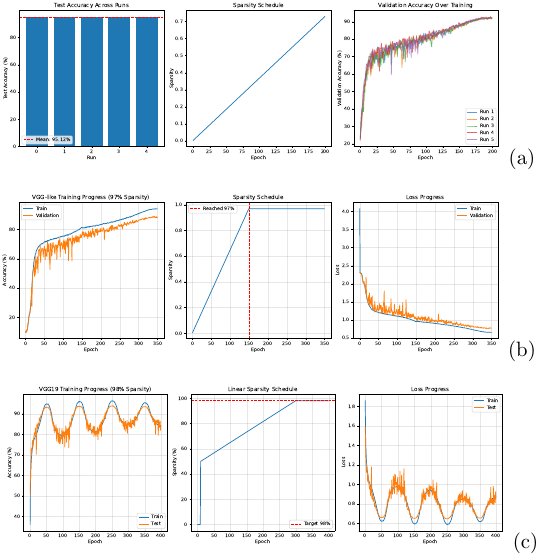}
    \caption{Training dynamics for progressive magnitude-based pruning on CIFAR-10 under the three baseline comparison settings. Each row reports the training and validation accuracy, sparsity schedule, and loss evolution for one architecture-sparsity configuration.}
    \label{fig:cifar10_baseline_training}
\end{figure*}

Figure~\ref{fig:cifar10_baseline_training} shows the corresponding training dynamics. Across the three architectures, the learning curves indicate stable convergence despite the progressive increase in sparsity.

\subsubsection{MNIST Baseline Comparisons}

Table~\ref{tab:mnist_comparison} presents the MNIST baseline comparisons using LeNet-300-100. These experiments evaluate whether the proposed progressive magnitude-based pruning method remains effective on a fully connected architecture and a simpler image classification benchmark.

\begin{table}[ht!]
	\centering
	\caption{MNIST performance comparison on LeNet-300-100 under aligned architecture and sparsity settings.}
	\label{tab:mnist_comparison}
	\begin{tabular}{lcc}
		\toprule
		\textbf{Method}               & \textbf{Sparsity (\%)} & \textbf{Accuracy (\%)} \\ \midrule
		LTH~\cite{frankle2019lottery} &          78.9          &       $\sim$98.5       \\
		\textbf{Ours}                 &          78.9          &     \textbf{98.88}     \\ \midrule
		SNIP~\cite{snip}              &          98.0          &          97.6          \\
		\textbf{Ours}                 &          98.0          &     \textbf{98.35}     \\ \bottomrule
	\end{tabular}
\end{table}

The proposed method outperforms both baselines under the corresponding sparsity settings. At 78.9\% sparsity, it improves over LTH by approximately 0.38 percentage points while using a single training cycle rather than iterative pruning and retraining. At 98.0\% sparsity, it surpasses SNIP by 0.75 percentage points, showing that progressive pruning remains effective even when only 2\% of the connections are retained.

These results indicate that the proposed pruning strategy generalizes beyond convolutional architectures. In particular, its effectiveness on LeNet-300-100 suggests that progressive magnitude-based adaptation can preserve important connections in fully connected networks as well as in convolutional models.

\begin{figure*}[ht!]
    \centering
    \includegraphics[width=0.95\linewidth]{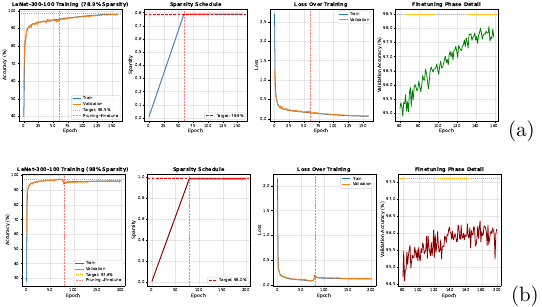}
    \caption{Training dynamics for progressive magnitude-based pruning on MNIST using LeNet-300-100 under the two baseline comparison settings. Each row shows the training and validation accuracy, sparsity schedule, loss evolution, and fine-tuning phase detail for one sparsity configuration.}
    \label{fig:mnist_baseline_training}
\end{figure*}

Figure~\ref{fig:mnist_baseline_training} shows the corresponding training dynamics. The curves illustrate stable convergence under both sparsity settings and show the transition from progressive pruning to fine-tuning at the target sparsity level.

\subsection{Ablation Study}

To isolate the contribution of the main design choices in the proposed method, we conduct an ablation study on ResNet-18 using CIFAR-10 at 72.9\% sparsity. The study compares the proposed magnitude-based pruning strategy with random pruning and examines the effects of the sparsity schedule, warm-up, and fine-tuning stages. The results are reported in Table~\ref{tab:ablation_study}.

\begin{table*}[ht!]
\centering
\caption{Ablation study on ResNet-18 using CIFAR-10 at 72.9\% sparsity. All configurations use a single training cycle.}
\label{tab:ablation_study}
\small
\begin{tabular}{lcc}
\toprule
\textbf{Configuration} & \textbf{Accuracy (\%)} & \textbf{Complexity} \\ \midrule
Random pruning (no magnitude) & 92.35 & Simple \\
\textbf{Ours: Magnitude + Linear, no warm-up, no fine-tuning} & \textbf{95.08} & \textbf{Simple} \\
Magnitude + Linear + Fine-tuning & 95.19 & +fine-tuning stage \\
Magnitude + Linear + Warm-up & 95.30 & +warm-up stage \\
Magnitude + Linear + Warm-up + Fine-tuning & 95.17 & +warm-up and fine-tuning stages \\
Magnitude + Polynomial schedule & 95.31 & Complex schedule \\ \bottomrule
\end{tabular}
\end{table*}

Table~\ref{tab:ablation_study} presents the ablation study on ResNet-18 at 72.9\% sparsity. Replacing magnitude-based pruning with random pruning results in a 2.73 percentage point accuracy drop (92.35\% vs. 95.08\%), confirming that the magnitude criterion is the most critical design choice. The polynomial schedule achieves marginally higher accuracy (95.31\%) compared with the linear schedule (95.08\%), but we adopt the linear schedule for its simplicity and ease of interpretation. Adding warm-up (95.30\%) or fine-tuning (95.19\%) individually provides marginal improvements of 0.22 and 0.11 percentage points, respectively. Combining both warm-up and fine-tuning yields 95.17\%, which is slightly lower than either alone, suggesting that these phases do not compound in this setting. Despite not employing warm-up or fine-tuning, techniques known to improve accuracy in the pruning literature, our baseline method remains competitive, demonstrating that the proposed magnitude-based progressive pruning achieves strong performance through simplicity rather than complex training procedures.

\subsection{Sparsity-Accuracy Trade-off Analysis}

To characterize the behavior of the proposed method across different compression levels, we conduct a sparsity sweep on ResNet-18 using CIFAR-10. The target sparsity levels range from 70\% to 95\%, and each experiment is repeated using five random seeds.

\subsubsection{Comprehensive Sparsity Sweep}

Table~\ref{tab:sparsity_tradeoff} reports the accuracy, remaining parameter ratio, and accuracy drop relative to the dense baseline.

\begin{table}[ht!]
	\centering
	\caption{Sparsity-accuracy trade-off on ResNet-18 using CIFAR-10. Results are averaged over five random seeds.}
	\label{tab:sparsity_tradeoff}
	\begin{tabular}{cccc}
		\toprule
		\textbf{Sparsity (\%)} & \textbf{Accuracy (\%)} & \textbf{Params Remaining} & \textbf{Accuracy Drop} \\ \midrule
		      0 (Dense)        &         95.27          &           100\%           &          ---           \\
		          70           &     95.21$\pm$0.10     &           30\%            &       $-$0.06pp        \\
		          75           &     95.22$\pm$0.18     &           25\%            &       $-$0.05pp        \\
		          80           &     95.24$\pm$0.06     &           20\%            &       $-$0.03pp        \\
		          85           &     95.18$\pm$0.15     &           15\%            &       $-$0.09pp        \\
		          90           &     95.06$\pm$0.19     &           10\%            &       $-$0.21pp        \\
		          95           &     93.86$\pm$0.09     &            5\%            &       $-$1.41pp        \\ \bottomrule
	\end{tabular}
\end{table}

The results show that accuracy remains close to the dense baseline up to 90\% sparsity. Between 70\% and 85\% sparsity, the accuracy remains within 0.1 percentage points of the dense model, indicating that a large fraction of ResNet-18 parameters can be removed without measurable performance degradation. At 90\% sparsity, the model retains 95.06\% accuracy, corresponding to a 0.21 percentage point drop.

A clearer degradation appears at 95\% sparsity, where accuracy decreases to 93.86\%. This suggests that, for ResNet-18 on CIFAR-10, the practical operating range of the proposed method is approximately 70--90\% sparsity, while 95\% represents a more aggressive compression regime with a noticeable accuracy trade-off.

\begin{figure}[ht!]
    \centering
    \includegraphics[width=0.95\linewidth]{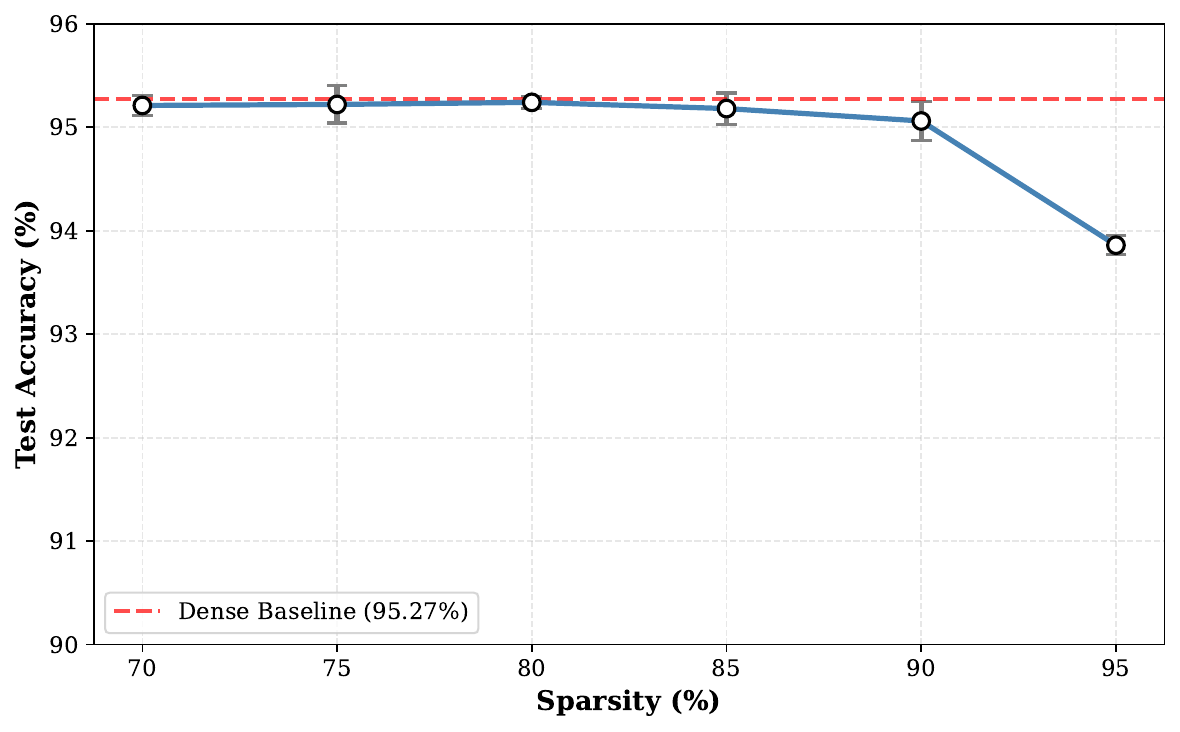}
    \caption{Sparsity-accuracy trade-off for ResNet-18 on CIFAR-10. Points show mean accuracy over five seeds, with error bars indicating $\pm$1 standard deviation.}
    \label{fig:sparsity_accuracy_curve}
\end{figure}

Figure~\ref{fig:sparsity_accuracy_curve} visualizes the same trend, including the standard deviation across the five random seeds.

\subsubsection{Progressive Sparsification Dynamics}

\begin{figure*}[ht!]
\centering
\includegraphics[width=0.95\linewidth]{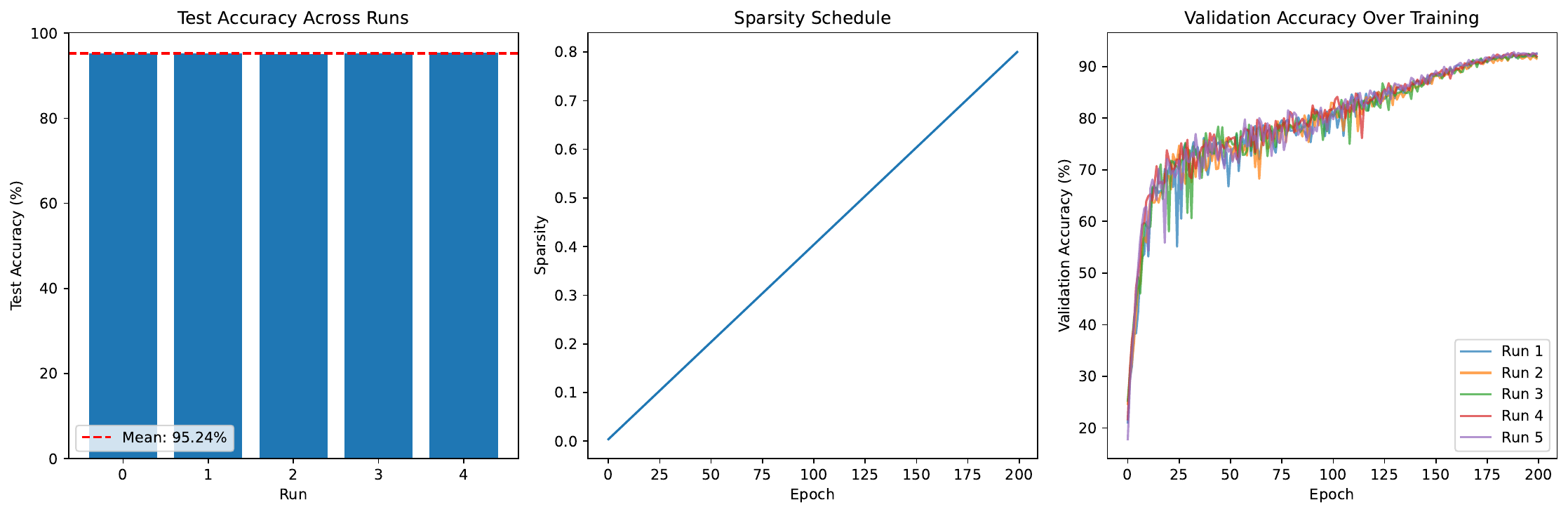}
\caption{Training dynamics for ResNet-18 on CIFAR-10 at 80\% target sparsity across five random seeds. The figure shows the final test accuracy distribution, sparsity schedule, and validation accuracy evolution.}
\label{fig:sparsity_evolution}
\end{figure*}

To examine the training behavior under progressive pruning, we analyze the 80\% sparsity configuration. Figure~\ref{fig:sparsity_evolution} shows the final accuracy distribution, sparsity schedule, and validation accuracy curves across five random seeds.

The validation curves indicate stable learning throughout the sparsification process. During early training, accuracy increases rapidly while sparsity remains relatively low. As sparsity increases, validation accuracy continues to improve, suggesting that the pruning process removes redundant connections without disrupting optimization. After the target sparsity is reached, continued training allows the remaining weights to stabilize, leading to consistent final performance across seeds.

Overall, the sparsity sweep and training dynamics indicate that progressive magnitude-based pruning maintains dense-level accuracy over a broad compression range and degrades gradually only at very high sparsity.

\subsection{Comparison with Dense Baseline}

Table~\ref{tab:dense_comparison} compares the proposed method with a dense ResNet-18 baseline on CIFAR-10 under the same training configuration.

\begin{table}[htbp]
\centering
\caption{Comparison with the dense ResNet-18 baseline on CIFAR-10.}
\label{tab:dense_comparison}
\begin{tabular}{lcccc}
\toprule
\textbf{Dataset} & \textbf{Architecture} & \textbf{Method} & \textbf{Sparsity} & \textbf{Accuracy (\%)} \\
\midrule
\multirow{2}{*}{CIFAR-10} & \multirow{2}{*}{ResNet-18} & Dense & 0\% & 95.27 \\
& & \textbf{Ours} & 80\% & 95.24$\pm$0.06 \\
\bottomrule
\end{tabular}
\end{table}

At 80\% sparsity, the proposed method achieves 95.24\% accuracy, compared with 95.27\% for the dense baseline. This corresponds to a 0.03 percentage point decrease while retaining only 20\% of the parameters, equivalent to a $5\times$ reduction in parameter count. This suggests substantial parameter redundancy in ResNet-18 for CIFAR-10 under the evaluated setting.

This result indicates that progressive magnitude-based pruning can substantially reduce model size while preserving dense-level accuracy on CIFAR-10, with potential efficiency benefits when sparse computation is used.

\section{Discussion}

The experimental results show that progressive magnitude-based pruning can produce highly sparse networks while preserving competitive accuracy on CIFAR-10 and MNIST. Across the evaluated architectures, the proposed method matches or exceeds the reported performance of representative iterative and initialization-based pruning methods, while using a single training cycle.

\subsection{Main Findings}

The CIFAR-10 results demonstrate that the proposed method performs well across different convolutional architectures and sparsity levels. On ResNet-18 at 72.9\% sparsity, it achieves 95.12\% accuracy compared with the reported LTH accuracy of 90.5\%. At higher sparsity levels, it also outperforms SNIP on the VGG-like architecture at 97\% sparsity and GraSP on VGG-19 at approximately 98\% sparsity.

The MNIST results further show that the method generalizes to fully connected architectures. On LeNet-300-100, it achieves 98.88\% accuracy at 78.9\% sparsity and 98.35\% accuracy at 98\% sparsity, exceeding the corresponding reported LTH and SNIP baselines.

The sparsity sweep on ResNet-18 provides additional insight into the accuracy-compression trade-off. Accuracy remains within 0.1 percentage points of the dense baseline for sparsity levels between 70\% and 85\%, and the model still achieves 95.06\% accuracy at 90\% sparsity. A clearer degradation appears at 95\% sparsity, where accuracy decreases to 93.86\%. These results suggest that 70--90\% sparsity is a practical operating range for ResNet-18 on CIFAR-10 under the evaluated configuration.

\subsection{Comparison with Existing Pruning Paradigms}

The proposed method occupies an intermediate position between iterative pruning, initialization-based pruning, and dynamic sparse training. Unlike iterative approaches such as LTH, it does not require repeated cycles of training, pruning, reinitialization, and retraining. Instead, it progressively updates the sparse structure during a single training run.

Compared with initialization-based methods such as SNIP and GraSP, the proposed method does not commit to a fixed sparse structure before learning begins. Its pruning decisions are based on weight magnitudes that evolve during training, allowing the sparse mask to reflect learned weight importance rather than only initialization-time criteria.

Compared with dynamic sparse training methods such as SET and RigL, the proposed method uses monotonic pruning without connection regrowth. This reduces implementation complexity because it avoids explicit regrowth criteria and candidate-connection management. Although this limits exploration of alternative sparse topologies, the results indicate that gradual monotonic pruning is sufficient to obtain strong performance in the evaluated settings.

\subsection{Interpretation of the Proposed Method}

The effectiveness of progressive magnitude-based pruning can be attributed to the interaction between optimization and gradual sparsification. During training, weights that contribute more strongly to the loss reduction tend to develop larger magnitudes, while less useful weights remain small. By progressively removing low-magnitude weights, the method allows the network to adapt to increasing sparsity without abrupt structural disruption.

The linear schedule also contributes to training stability. Because sparsity increases gradually, the model has time to adjust its remaining weights before additional connections are removed. This is particularly important at high sparsity levels, where the remaining connectivity becomes increasingly constrained.

The use of a magnitude-based criterion is also computationally simple. It does not require additional gradient, Hessian, or saliency computations beyond standard training. The results therefore suggest that, when applied progressively, a simple magnitude criterion can remain competitive with more complex initialization-based pruning criteria.

\subsection{Design Considerations}

Several implementation choices support the coherence and stability of the proposed pruning procedure. Thresholds are computed using only currently active weights, preventing already-pruned zero weights from affecting later pruning decisions. The no-regrowth policy keeps the sparse structure monotonic and avoids explicit connection-regrowth mechanisms. Finally, mask enforcement before the forward pass, during gradient updates, and after optimizer updates ensures that pruned weights remain inactive throughout training, including when momentum-based optimization is used.

Together, these choices define a consistent progressive pruning procedure that maintains strict sparsity while allowing the remaining weights to adapt during training.

\subsection{Limitations}

This study focuses on image classification using CIFAR-10 and MNIST with ResNet, VGG-style, and LeNet architectures. Although these benchmarks allow direct comparison with established pruning baselines, the generality of the method remains to be tested on larger datasets, such as ImageNet, and on other architectures, such as transformers and graph neural networks.

The method also uses global unstructured pruning. While this achieves high parameter sparsity, it does not necessarily translate into practical inference speedup on standard hardware unless sparse computation is efficiently supported. Structured pruning, which removes channels, filters, or layers, may provide more direct deployment benefits.

Finally, the pruning criterion is based solely on weight magnitude. Although this choice is efficient and effective in the reported experiments, it may not capture all aspects of weight importance in more complex architectures or tasks.

\subsection{Future Work}

Future work should evaluate the proposed method on larger-scale datasets and modern architectures, including ImageNet-scale vision models and transformer-based models. Extending the framework to structured pruning is also an important direction, since structured sparsity is more directly compatible with standard hardware acceleration.

Another promising direction is to combine progressive pruning with complementary compression techniques, such as quantization or knowledge distillation. Finally, lightweight hybrid criteria that combine magnitude with gradient or structural information may improve pruning decisions while preserving the simplicity of the proposed approach.

\section{Conclusion}

This work evaluated progressive magnitude-based pruning as a practical approach for neural network sparsification. The proposed method uses a linear sparsity schedule with per-epoch mask updates, allowing sparse connectivity patterns to emerge progressively during training while avoiding iterative retraining and connection-regrowth mechanisms.

The experimental results on CIFAR-10 and MNIST show that the proposed method achieves competitive or superior performance compared with representative pruning baselines across different paradigms. On CIFAR-10, it outperforms LTH on ResNet-18 at 72.9\% sparsity, SNIP on a VGG-like architecture at 97\% sparsity, and GraSP on VGG-19 at approximately 98\% sparsity. On MNIST, it also exceeds the reported LTH and SNIP results on LeNet-300-100, demonstrating that the method is effective for both convolutional and fully connected architectures.

The sparsity-accuracy analysis further shows that ResNet-18 maintains near-dense accuracy across a broad sparsity range. In particular, accuracy remains within 0.1 percentage points of the dense baseline between 70\% and 85\% sparsity, and remains highly competitive at 90\% sparsity. These findings suggest that progressive magnitude-based pruning can provide substantial parameter reduction with limited accuracy degradation under the evaluated settings.

Overall, the results indicate that progressive magnitude-based pruning offers a balanced alternative to iterative, initialization-based, and dynamic sparse training approaches. It combines single-cycle training, adaptive mask evolution, and a controlled monotonic pruning procedure, making it a practical baseline for sparse neural network training under the evaluated settings.

\bibliographystyle{ieeetr}
\bibliography{references}

\end{document}